# Leveraging GPT-4 for Food Effect Summarization to Enhance Product-Specific Guidance Development via Iterative Prompting

Yiwen Shi[1], Ping Ren[2], Jing Wang[2], Biao Han[2], Taha ValizadehAslani[3], Felix Agbavor[4], Yi Zhang[2], Meng Hu[2], Liang Zhao[2], Hualou Liang[4*]

[1] College of Computing and Informatics, Drexel University, Philadelphia, PA, United States
[2] Office of Research and Standards, Office of Generic Drugs, Center for Drug Evaluation and Research, United States Food and Drug Administration, Silver Spring, MD, United States
[3] Department of Electrical and Computer Engineering, College of Engineering, Drexel University, Philadelphia, PA, United States
[4] School of Biomedical Engineering, Science and Health Systems, Drexel University, Philadelphia, PA, United States

*Corresponding author: hualou.liang@drexel.edu

## Abstract

Food effect summarization from New Drug Application (NDA) is an essential component of product-specific guidance (PSG) development and assessment, which provides the basis of recommendations for fasting and fed bioequivalence studies to guide the pharmaceutical industry for developing generic drug products. However, manual summarization of food effect from extensive drug application review documents is time-consuming. Therefore, there is a need to develop automated methods to generate food effect summary. Recent advances in natural language processing (NLP), particularly large language models (LLMs) such as ChatGPT and GPT-4, have demonstrated great potential in improving the effectiveness of automated text summarization, but its ability with regard to the accuracy in summarizing food effect for PSG assessment remains unclear. In this study, we introduce a simple yet effective approach, *iterative prompting*, which allows one to interact with ChatGPT or GPT-4 more effectively and efficiently through multi-turn interaction. Specifically, we propose a three-turn iterative prompting approach to food effect summarization in which the keyword-focused and length-controlled prompts are respectively provided in consecutive turns to refine the quality of the generated summary. We conduct a series of extensive evaluations, ranging from automated metrics to FDA professionals and even evaluation by GPT-4, on 100 NDA review documents selected over the past five years. We observe that the summary quality is progressively improved throughout the iterative prompting process. Moreover, we find that GPT-4 performs better than ChatGPT, as evaluated by FDA professionals (43% vs. 12%) and GPT-4 (64% vs. 35%). Importantly, all the FDA professionals unanimously rated that 85% of the summaries generated by GPT-4 are factually consistent with the golden reference summary, a finding further supported by GPT-4 rating of 72% consistency. Taken together, these results strongly suggest a great potential for GPT-4 to draft food effect summaries that could be reviewed by FDA professionals, thereby improving the efficiency of PSG assessment cycle and promoting the generic drug product development.

# 1. Introduction

The U.S. Food and Drug Administration (FDA) publishes product-specific guidances[1] (PSGs) to outline the agency's current thinking on how to establish the bioequivalence between a proposed test drug product and the corresponding reference standard drug product. The PSGs guide and facilitate generic drug product development, accelerate Abbreviated New Drug Application (ANDA) submission and approval, and ultimately ensure public access to safe, effective, high-quality, and affordable generic drugs. However, the PSG assessment process can be labor-intensive, especially in collecting supportive information. As a result, there is a growing need to automate information retrieval from different data sources (Shi et al., 2021). Automating the collection of pharmacokinetics information, such as absorption, distribution, metabolism, and excretion (ADME) of Reference Listed Drug (RLD) products, is a critical step in the PSG assessment enhancement (Shi et al., 2023). A major challenge in reducing the burden of PSG assessment is the automatic summarization of key elements from lengthy documents. For example, the food effect is an essential aspect of drug absorption, as it can influence pharmacokinetics through various mechanisms, such as delaying gastric emptying, changing the pH of the gastrointestinal tract, and physically or chemically interacting with the dosage form or drug (Sharma et al., 2013). In the RLD product application review documents, food effect studies demonstrate whether there is a clinically significant impact of food on the product's bioavailability or pharmacokinetics. However, these documents are typically dozens of pages in length and can be time-consuming to summarize manually. To address this challenge, it is desirable to automatically generate an initial version of the summary, from which the FDA staff can refine to ensure that it accurately reflects the relevant information.

Large Language Models (LLMs) have demonstrated exceptional performance in zero/few-shot tasks in NLP (Bai et al., 2022; Brown et al., 2020), which also raise interest in their potential for automatic text summarization (Goyal et al., 2022; Liu et al., 2022). AI-powered chatbots, ChatGPT (OpenAI, 2022) and GPT-4 (OpenAI, 2023), represent the latest generation in LLMs. Although these models are not originally developed for the biomedical domain, their capabilities have shown significant promise in biomedical research (Gao et al., 2023; Holmes et al., 2023; Lee et al., 2023).

Prior research has yet to explore the potential of leveraging the outputs of LLMs to refine text summaries iteratively. In this study, we propose a simple strategy, *iterative prompting*, which allows the iterative conversation with the chatbot. Specifically, we leverage two state-of-the-art chatbots, ChatGPT and GPT-4, to generate summaries of food effect studies in NDA review files. We formulate the iterative conversation between humans and chatbots while considering its potential to improve the accuracy and efficiency of food effect study summarization. Figure 1 shows our proposed approach with an example of an iterative conversation with ChatGPT. The transcript highlights how ChatGPT enhances the summary via collaborating with human.

---

[1] https://www.accessdata.fda.gov/scripts/cder/psg/index.cfm

**Summarize the following text related to food effect studies.**

*EMR200095-002: crossover study to investigate the food effect and relative BA of tepotinib in 2 groups of healthy subjects.*

*MS200095-0044: study with crossover design in each part to investigate the BE of the tablet formulation of tepotinib TF3 compared to TF2, and to investigate the influence of food on the PK of each tablet formulation TF2 and TF3 of tepotinib in healthy subjects.*

*The recommendation that tepotinib is administered with food is supported by the food effect study showing the mean AUC0-INF of tepotinib increased by 1.6-fold and Cmax increased by 2-fold (Table 18), following administration of a high-fat, high-calorie meal (approximately 800 to 1,000 calories, 150 calories from protein, 250 calories from carbohydrate, and 500 to 600 calories from fat). The median Tmax shifted from 12 hours to 8 hours. In addition, taking tepotinib with food is expected to mitigate the gastrointestinal toxicities including diarrhea, which were among the most common adverse reactions. The food effect is considered clinically meaningful, as a reduction in exposure following administration of tepotinib without food may potentially reduce anti-tumor activity and tepotinib was administered with food in VISION. ……*

The food effect study shows that administering tepotinib with food increases its exposure and mitigates adverse reactions like diarrhea. The study also indicates that taking tepotinib without food may reduce its anti-tumor activity. The recommended mode of administration *for* tepotinib is with food. The crossover studies investigate the food effect and relative bioavailability of tepotinib in healthy subjects and compare the BE of the different formulations of tepotinib. FDA recommends administering tepotinib with food.

**Add facts of AUC, Cmax, and Tmax in summary.**

In the food effect study, it was found that taking tepotinib with a high-fat, high-calorie meal increased the mean AUC0-INF by 1.6-fold and Cmax by 2-fold. The median Tmax also shifted from 12 hours to 8 hours. Similarly, the crossover studies showed an increase in AUC and Cmax when tepotinib was administered with food. Thus, taking tepotinib with food is recommended to enhance its exposure and clinical outcomes.

**Summarize it in 2-3 sentences and keep the detail of AUC, Cmax, and Tmax.**

The food effect study found that taking tepotinib with a high-fat, high-calorie meal increased the mean AUC0-INF by 1.6-fold and Cmax by 2-fold, with a shift in median Tmax from 12 hours to 8 hours. The recommended mode of administration for tepotinib is with food to enhance its exposure and clinical outcomes.

Figure 1: An example to illustrate the iterative conversation with the ChatGPT for text summarization of food effect study (NDA214096) used in our approach. The conversation shows how the summary is improved progressively via interactive chats that are akin to multi-turn prompt engineering.

We evaluate our proposed method with a comprehensive assessment including both human and automatic metric evaluations. Recent advancements in GPT-4 have demonstrated its ability to produce highly consistent rankings and perform detailed assessments when comparing chatbot responses (Chiang et al., 2023), and even identify errors made by humans and AI (Lee et al., 2023). In this study, we set out to investigate GPT-4's evaluation capabilities, which could potentially enable an automated evaluation framework for summarization assessments.

The contributions of our work are as follows:

- We propose a simple yet effective strategy, *iterative prompting*, to allow one to interact with ChatGPT or GPT-4 more effectively and efficiently through multi-turn interaction, which iteratively improves the quality of text summary.

- We carry out extensive evaluations of our approach using 100 NDA review documents for food effect summarization and demonstrate the potential of employing multi-turn interaction to refine the quality of the generated summaries.
- We demonstrate the superior performance of GPT-4 over ChatGPT and highlight the potential of GPT-4 in automating evaluation while acknowledging certain limitations.

## 2. Related Work

**Text Summarization:** Text summarization is the task of generating a concise paragraph that captures the key points of an article, offering benefits such as time-saving and helping individuals identify relevant information (Lewis et al., 2019; See et al., 2017; Zhang et al., 2019). The introduction of sequence-to-sequence models and attention mechanisms (Sutskever et al., 2014) has resulted in rapid progress on extractive (Nallapati et al., 2017), abstractive (Nallapati et al., 2016; Zhang et al., 2019), and hybrid models (Gu et al., 2016; See et al., 2017) for summarization. In the biomedical field, these methods have primarily been employed for summarizing research publications or electronic health record (EHR) data (Chaves et al., 2022; Mishra et al., 2014) and clinical trials (Cintas et al., 2019). In this study, we present a case study focused on summarizing text related to food effect studies from drug review files, aiming to facilitate PSG assessment.

The automatic evaluation of text summarization often relies on comparisons with the gold standard, and previous studies have indicated that these metrics are ineffective for evaluating summaries, particularly when using LLMs (Deutsch et al., 2022; Fabbri et al., 2021; Zhang et al., 2023)

**Large Language Models:** The LLMs have emerged as a notable development in NLP, drawing inspiration from previous pretrained models (Brown et al., 2020; Devlin et al., 2019; Radford et al., 2019) while operating at significantly larger scales and introducing novel advancements (Google, 2022; Meta, 2023; OpenAI, 2023). These LLMs are built upon the transformer architecture (Vaswani et al., 2017), which exhibits a substantial increase in scale in terms of model parameters and training data, enabling them to capture a more comprehensive understanding of language. Additionally, LLMs possess the remarkable capability of performing zero-shot or few-shot learning without the need for fine-tuning (Kojima et al., 2023; Wei et al., 2023).

In addition to the significant progress made in model architecture, scale, and training strategies, LLMs can be further aligned with human preferences through reinforcement learning from human feedback (RLHF, Christiano et al., 2023). This approach has been implemented in various LLMs, including ChatGPT. By incorporating RLHF, ChatGPT emphasizes adhering to prompts and generating comprehensive responses, enabling highly successful human-like interactions, which makes it an ideal candidate for this study.

The development of GPT-4 has significantly pushed the boundaries of state-of-the-art language models, which showcases the enhancement in reasoning abilities, image comprehension, multi-modal capabilities, resulting in more sophisticated and diverse responses (OpenAI, 2023; Zheng et al., 2023).

**Prompt Engineering:** Prompt engineering is the process of designing or crafting effective prompts to elicit desired responses from language models like ChatGPT. Prompting or in-context learning enables the possibility of adaptation of pre-trained language models to downstream tasks without fine-tuning (Brown et al., 2020). A prompt serves as a set of instructions to customize the response of the language model. The advent of prompt engineering signifies a new era in NLP (Liu et al., 2021). Prompts can be generated manually or automatically (Jiang et al., 2020; Schick & Schütze, 2021). Although automatically generated prompts may outperform manual prompts in specific tasks, they can suffer from issues related to human readability (Taylor et al., 2022). Consequently, manual prompt generation may be preferred in domains where interpretability is essential. In this study, we design prompts based on the interactive feature of ChatGPT, which enables human collaboration with the underlying LLM to improve its performance.

## 3. Our Approach: GPT-4 for Food Effect Summarization via Iterative Prompting

AI-powered chatbots like ChatGPT and GPT-4 are designed to interact with humans conversationally. The dialogical format allows the model to remember previous outputs in its "working memory", enabling it to answer follow-up questions and adjust its output based on human feedback. For text summarization of the food effect study, we propose a three-turn interactive conversation between a PSG assessor and a chatbot (shown in Figure 2). Each iteration contains a query or constraint in the prompt reflecting the human feedback based on the previous output.

In the first turn, we define the task's goal: *Summarize the following text related to food effect studies.* The model typically generates a high-level summary without including the details of the pharmacokinetic metrics of the food effect study. In the second turn, we focus on keyword-focused summarization to generate coherent summaries containing key pharmacokinetic metrics, such as AUC, Cmax, and Tmax, in the food effect study. In the final turn, we use a sentence-count length prompt to ensure the summary is concise, making it easier for the PSG assessor to locate the key points. We determine that the food effect section in the drug labeling provides an ideal length for the food effect study summary, with approximately 90% of them within three sentences.

> **Turn 1 (task instruction):**
> *Summarize the following text related to food effect studies. {article}*
> **Turn 2 (keyword-constrained prompt):**
> *Add facts of AUC, Cmax, and Tmax in the summary.*
> **Turn 3 (length-constrained prompt):**
> *Summarize it in 2-3 sentences and keep the detail of AUC, Cmax, and Tmax.*

Figure 2: Our proposed three-turn iterative prompting approach that enables human collaboration with the underlying chatbot via interactive chats to refine the quality of food effect summarization.

Overall, this approach represents a pipeline for continuously improving the quality of responses and enhancing communication with the chatbot in text summarization tasks.

## 4. Experimental Setup

### 4.1 Dataset

The dataset for this study consists of two parts: the detailed food effect study documents from NDA review files available to the public via Drugs@FDA[2] website, which serve as the article to be summarized, and the corresponding concise food effect section from drug labeling, serving as the ground-truth reference summary.

The FDA NDA review files comprise a compilation of various documents, including multi-discipline reviews, integrated reviews, clinical pharmacology reviews, and others. These review files contain comprehensive information that is crucial for the FDA assessment of the safety and efficacy of drugs, including the study of food effects. A team of the FDA professionals or FDA assessors (PR, MH, JW, BH) selected new drug applications from the past five years (from 2019 to the present). We excluded the drugs that lacked publicly available drug reviews or a food effect section in the drug labeling and ended up with 100 drugs used for this study. These drugs provide a comprehensive coverage of the NDA drugs in terms of clinical category, drug substance and dosage form. A summary of the drug products is shown in the Appendix A.

To accommodate the prompt size limitation of ChatGPT, the FDA assessors thoroughly reviewed the NDA review files, annotating the sections related to the food effect studies as the articles to be summarized. To create the corresponding golden reference summaries, we manually collect the food effect section from the FDA-approved drug labeling available on DailyMed[3].

### 4.2 Models Used

In this study, we utilize two LLMs, ChatGPT and GPT4, to generate summaries of food effect studies. ChatGPT is an improved version of the highly acclaimed GPT-3 model (Brown et al., 2020), which has 175 billion parameters and was trained on extensive data collection, including Wikipedia. It can learn from task instruction or a few demonstrative examples in context without updating model parameters (Ouyang et al., 2022). ChatGPT inherits these features from GPT-3 and uses RLHF (Christiano et al., 2023), which enhances its capabilities for text generation, language understanding, and natural language processing. On the other hand, GPT-4 is the latest iteration in the GPT series and is the most powerful LLM to date, offering even more advanced features and capabilities than its predecessors. Although OpenAI has yet to disclose many technical details about GPT-4, it has been shown to outperform ChatGPT in various tasks (OpenAI, 2023). In our implementation, we utilize `gpt-3.5-turbo` and `gpt-4` respectively for ChatGPT and GPT-4 through the OpenAI chat completion API[4].

### 4.3 Evaluation Protocol

We undertake a comprehensive assessment of ChatGPT and GPT-4's text summarization capabilities for the food effect study, aiming to address three key questions. First, whether the multi-turn iterative conversations with a chatbot based on human feedback can enhance summary quality. Second, which model (ChatGPT vs GPT-4) performs better for this particular task in the study? Third, since GPT-4 is the latest iteration in the GPT series, offering more advanced features and capabilities than its predecessors, we want to assess to what extent is GPT-4 able to generate a summary factually consistent with the reference summary from the drug labeling.

---

[2] https://www.accessdata.fda.gov/scripts/cder/daf/index.cfm
[3] https://dailymed.nlm.nih.gov/dailymed
[4] https://openai.com/blog/openai-api

### 4.3.1 Evaluation Tasks

We design three evaluation tasks to address these questions, each involving a human evaluation study. Additionally, we employ GPT-4 in our evaluation tasks as previous research has indicated its potential capability to automate chatbot assessment (Chiang et al., 2023). We also compute reference-based automated metrics using Recall-Oriented Understudy for Gisting Evaluation (ROUGE) (Lin, 2004), given that the drug labeling provides high-quality and concise summarization of the drug's food effect studies.

**Task 1: Comparison between Summaries of Different Iterations**

To evaluate if iterative prompt ChatGPT based on human feedback can enhance summary quality, we devise an evaluation task to collect preference annotations. For each given drug, assessors are shown summaries from all three turns in a randomized order to avoid bias in the order of presentation of the summaries. Assessors are required to choose their most and least preferred summary or summaries and provide an optional free-text justification.

**Task 2: Comparison between ChatGPT and GPT-4**

We conduct a blinded pairwise comparison evaluation for the summarization capability between ChatGPT and GPT-4. Assessors are presented with a set of summary pairs, each containing one summary generated by ChatGPT and another generated by GPT-4 in the last turn. The order of presentation of the summary pairs is randomized to minimize bias. Assessors are asked to choose the better summary within each pair or declare the summaries equally good if they cannot differentiate them.

**Task 3: GPT-4 Consistency Evaluation**

Despite significant improvements in reducing hallucinations compared to earlier GPT models, GPT-4 still has limitations and may produce unreliable summaries that hallucinate facts. We conduct this evaluation to assess whether GPT-4 generated summaries are factually consistent with the reference summary.

### 4.3.2 Evaluation Metrics

First, we assess the model performance using standard automated metrics of ROUGE. Second, we conduct human expert evaluation of food effect summaries produced by the models to mitigate some of the aforementioned limitations of the automated metrics. Third, we utilize GPT-4 to determine relevance of the model summary relative to the reference summary.

**Automated Metrics:** To evaluate the impact of iterative prompting and compare the performance of ChatGPT and GPT-4, we utilize automatic metrics ROUGE, which compares generated summaries against the reference summaries. Specifically, we compare the summary generated in each turn with the reference summary, to determine if the metrics can effectively evaluate the impact of iterative prompting and distinguish the performance between ChatGPT and GPT-4.

**Human Evaluation:** Three independent FDA assessors evaluate each task (PR, ML, JW). These FDA professionals all have at least three years of experience in the PSG assessment. We do not provide assessors with specific definitions of summary quality to avoid introducing biases.

Assessors rely on their own preferences based on summaries they would like to see for the food effect study.

**GPT-4 Evaluation:** Recent advancements in GPT-4 have shown its ability to produce highly consistent rankings and perform detailed assessments when comparing chatbots' responses (Chiang et al., 2023) and even identify errors made by humans and AI (Lee et al., 2023). This raises the question of whether GPT-4's evaluation capabilities can reach a level comparable to that of humans, potentially enabling an automated evaluation framework for summarization assessments. We employ GPT-4 in two evaluation tasks: comparing the performance of ChatGPT and GPT-4 and evaluating the factual consistency of GPT-4-generated summaries. The evaluation prompts used in these tasks are provided in the Appendix B and C.

## 5. Results

**Summaries from the last turn are most preferred.** Figure 3 shows the distribution of assessor votes for the most and least preferred summaries generated at each turn during the iterative prompting process. The results indicate that the summaries from Turn 3, which is the last turn, received the highest number of votes in "Most Preferred Summary" and are thus considered the most preferred summaries. Moreover, it is supported by all three assessors in almost half of the drugs. In contrast, the summaries generated in Turn 1, the first turn, received the most votes in the "Least Preferred Summary", indicating that they are the least preferred summaries[5]. However, it is worth noting that there are 42 drugs that receive zero votes for "Least Preferred Summary" in Turn 1, which suggests that ChatGPT is able to generate satisfactory summaries even with only task instructions.

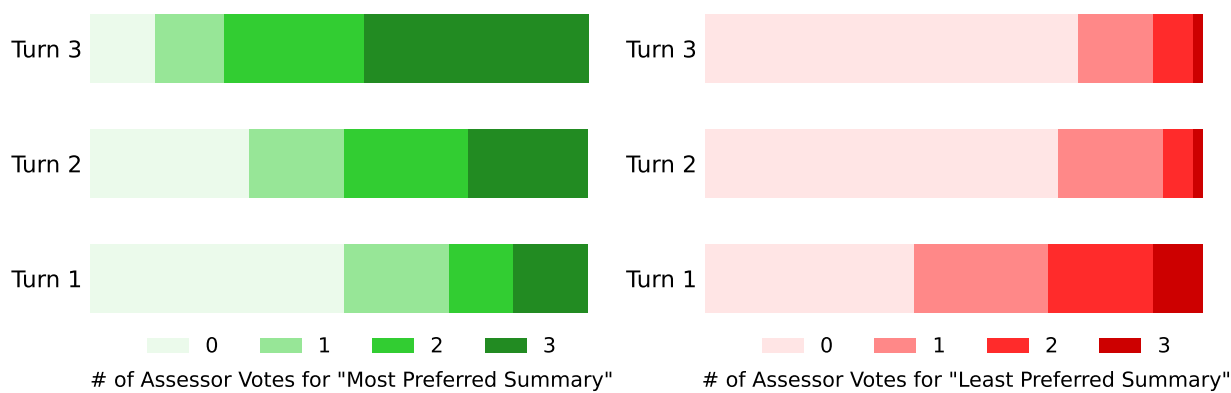

Figure 3: Distribution of assessor votes for the most and least preferred summaries at each turn during the iterative prompting process. Summaries from Turn 3 (the last turn) received the highest number of votes in the "Most Preferred Summary".

Additionally, we employed Krippendorff's alpha (Castro, 2017) to measure the inter-annotator agreement of the most and least preferred summaries. The results revealed moderate to substantial agreement among the assessors, with an alpha of 0.49 and 0.55 for the most and least preferred summary, respectively.

---
[5] Assessors may leave "Least Preferred Summary" empty when they think none of the summaries missed key content related to the food effect study.

Table 1: ROUGE scores (1/2/L) of summaries at each turn. Bold indicates the best score among the turns for each metric.

|  | ROUGE-1/2/L | |
|---|---|---|
|  | ChatGPT | GPT-4 |
| Turn1 | 29.40/11.24/19.48 | 31.44/12.72/20.45 |
| Turn2 | 32.67/**14.06**/21.75 | 31.37/**12.88**/20.98 |
| Turn3 | **34.04**/13.36/**22.60** | **33.64**/11.44/**22.12** |

Table 1 shows the results of the ROUGE-1/2/L evaluation of summaries generated at each turn. It indicates that the summary becomes increasingly similar to the reference summary as the iterative process progresses. Turn 3 outperforms other turns regarding ROUGE-1 and ROUGE-L scores for ChatGPT and GPT-4, which is consistent with the human evaluation results. It demonstrates that automatic reference-based ROUGE metrics effectively capture the improvement between turns.

**GPT-4 generated better summaries than ChatGPT.** The ROUGE scores in Table 1 indicated minimal differences for Turn 3 outputs between ChatGPT and GPT-4. As a result, human evaluation was relied upon for comparison. Figure 4 shows that human assessors and GPT-4 preferred the summaries generated by GPT-4 over ChatGPT. The majority vote of the human annotations[6] showed that GPT-4 won 43% of the time, while there was a tie between GPT-4 and ChatGPT 45% of the time. Human assessors reached a moderate agreement (Krippendorff's alpha is 0.35). However, the inter-annotator agreement between the human majority vote and GPT-4 is low (Krippendorff's alpha is 0.07), as GPT-4 struggled to distinguish scenarios where the summaries from both models were equally good.

---

[6] If all three assessors vote differently (e.g. one vote for "ChatGPT Won", "Tie" and "GPT-4 Won", respectively), the majority vote will be "Tie".

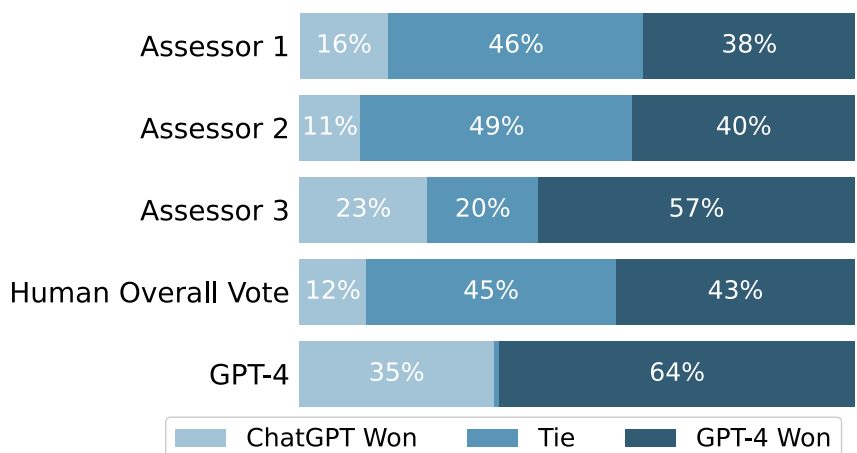

Figure 4: Human and GPT-4 evaluation results for comparing summaries generated by ChatGPT and GPT-4. Both human assessors and GPT-4 show strong preference of the summary generated by GPT-4 over ChatGPT. However, GPT-4 has difficulty identifying the scenario where the summaries are equally good.

**GPT-4 demonstrates the capability to generate summaries factually consistent with the reference summary.** Three assessors compare the GPT-4 generated summaries with the food effect section from drug labeling. The majority vote of the human annotations indicates that 85% of the GPT-4 generated summaries are factually consistent with the reference summary, with a reasonable agreement among assessors (Krippendorff's alpha is 0.41). However, GPT-4 evaluation shows a consistency rate of 72%, albeit low relative to human rating, there is rather strong consistency of rating between human and GPT-4 as both share 69% overlap.

## 6. Conclusion and Discussion

In this study, we propose a simple, yet effective strategy, iterative prompting, enabling human collaboration with the underlying LLM via interactive chats to refine the quality of food effect summarization. We carry out extensive evaluations of our approach using 100 NDA review documents for food effect summarization. By employing multi-turn interaction to refine the quality of the generated summaries, we show the great potential of GPT-4 as AI assistant to draft food effect summaries that could be reviewed and edited by human experts, thereby improving PSG assessment, and accelerating the generic drug product development.

The goal of text summarization is to extract essential information from documents and produce short, concise, and readable text. One main advantage of our approach over previous methods is that it interacts with ChatGPT in a conversational way. Specifically, it uses iterative prompting that enables human collaboration with the underlying LLM via interactive chats to refine the quality of food effect summarization. Prompt engineering is crucial in leveraging the potential of language models for generating natural language outputs such as summarization. However, it can be a delicate task as even slight prompt modifications can result in significant changes in model predictions, making it a fragile process. To address this challenge, researchers have proposed various methods for prompt optimization, such as chain-of-thought (Wei et al., 2022) and its variants (Wang et al., 2022; Yao et al., 2023) or even attaining a goal fully autonomously (Auto-GPT[7]). These techniques are quite expensive and usually also have trouble staying on task. We tackle this issue by leveraging the interactive feature of the chatbot that enables human feedback to steer the underlying LLM toward desirable or useful results pertinent to the food effect. It is the multi-turn interactivity that sets our method apart from existing approaches.

In this study, we used 100 NDA drug review documents for food effect summarization, which were sampled from the NDA drugs available over the past five years. Although the sample size is limited, these drugs nevertheless provide comprehensive coverage of the NDA drugs in terms of clinical category, drug substance and dosage form, as shown in the Appendix. In addition, the NDA review documents, which need to be summarized, vary in length from dozens of pages to several hundred pages. However, the input capacity of LLMs is restricted due to the constraint imposed on the length of prompts. For example, the prompt size (or the length of the context window) is limited to 4096 tokens for ChatGPT and 8,000 tokens (100 tokens are roughly 75 words) for GPT-4, which is currently accessible. To accommodate the token size limit, in this study we use the coarsely-labeled text pertaining to food effect annotated by the FDA assessors. The direct use of the full NDA review documents is preferred as it will further reduce the review time for text summarization. We leave it to future work.

**What kind of summaries do PSG assessors prefer?**

The assessors highlight three factors influencing the evaluation result. First, the summary must encompass all correct pharmacokinetic (PK) values (e.g., AUC, Cmax, and Tmax) in the food effect studies. Other than the three PK parameters that we have included in the keyword-constrain prompt, they are also interested in meal type (e.g., high/low-fat meal) and the clinical significance of the effect. Such insight offers helpful guidance to continue to refine the prompt in the future. Second, assessors may consider a penalty when irrelevant information is in the summary. It reveals a limitation of the current strategy: the quality of the input data directly impacts the output quality. If the article contains irrelevant information, such as drug-drug interactions, it may be reflected in the generated summary. For example, the article of NDA213246 contains a drug-drug interaction paragraph: "*Coadministration of a single 160 mg dose of selpercatinib capsules under fasted conditions with multiple daily doses of a PPI (omeprazole) decreased the AUC0-INF of selpercatinib by 69% and the Cmax by 88% as compared to selpercatinib administered alone*.", that could result in a summary that captures the PK values as "*Co-administration with omeprazole significantly decreased the AUC and Cmax, with a reduction of 69% and 88%, respectively, in the fasted state*." Last, assessors prefer more concise summaries if they include all the key metrics without irrelevant information.

Furthermore, we have observed that while the length constraint imposed on the summary effectively condenses the information, it also poses a potential drawback of omitting essential details of key metrics. Figure 5 provides an example where the PK metrics are presented in the summary after the second iteration as "*...with an increase of 22% in AUC and 42% in Cmax when a 345mg tablet is administered in the fed state...*". However, ChatGPT's rewritten form during the third iteration is expressed as "*slightly increased AUC and significantly increased Cmax*." Therefore, careful consideration is needed to strike the right balance between conciseness and the inclusion of essential information in the summary.

---

[7] https://github.com/Significant-Gravitas/Auto-GPT

> **Summarize the following text related to food effect studies.** *{article}*

**ChatGPT:** Elacestrant is recommended to be taken orally once daily, with food, at a dosage of 345 mg. Clinical studies in breast cancer patients have shown improved gastrointestinal tolerability when elacestrant is taken with food, and food intake has been found to increase its systemic exposure in the body. However, the difference in exposure between fasted and fed conditions is not considered clinically relevant, and administering elacestrant with food is recommended to minimize the risk of gastrointestinal adverse reactions. No therapeutic individualization is needed based on demographic factors, renal impairment, or mild hepatic impairment. However, dose reduction is necessary in patients with concomitant moderate hepatic impairment, and elacestrant should not be taken with strong or moderate CYP3A4 inhibitors and/or inducers.

**GPT-4:** Elacestrant is recommended to be taken at a dosage of 345 mg orally once daily with food. Clinical studies show improved gastrointestinal tolerability when administered with food. While there is a slight increase in exposure with high-fat meals compared to low-fat meals, it is not clinically relevant. The FDA advises taking elacestrant with food to minimize gastrointestinal adverse reactions. No therapeutic individualization is required based on demographic factors, renal impairment, or mild hepatic impairment. However, dose reduction is necessary for patients with concomitant moderate hepatic impairment, and concomitant use with strong or moderate CYP3A4 inhibitors and/or inducers should be avoided.

> **Add facts of AUC, Cmax, and Tmax in summary.**

**ChatGPT:** The presence of food, specifically a high-fat meal, has been found to increase the systemic exposure of elacestrant compared to fasted conditions, with an increase of 22% in AUC and 42% in Cmax when a 345mg tablet is administered in the fed state. Tmax is not specifically mentioned in the text. However, it is important to note that improved gastrointestinal tolerability was observed when elacestrant tablets were taken with food, despite the slightly higher but not clinically relevant exposure, as observed in the study.

**GPT-4:** When taken with food, elacestrant shows increased systemic exposure compared to fasted conditions. The area under the curve (AUC) increased by 22%, and the maximum concentration (Cmax) increased by 42%. The time to reach maximum concentration (Tmax) was not mentioned in the text provided. Taking elacestrant with food, particularly a high-fat meal, improves gastrointestinal tolerability and is recommended by the FDA to minimize the risk of gastrointestinal adverse reactions.

> **Summarize it in 2-3 sentences and keep the detail of AUC, Cmax, and Tmax.**

**ChatGPT:** Elacestrant should be taken orally once daily with food at a recommended dosage of 345mg to minimize gastrointestinal adverse reactions. Administration with a high-fat meal has been found to slightly increase AUC and significantly increase Cmax. Tmax is not mentioned, but improved gastrointestinal tolerability was observed when tablets were taken with food despite the increased exposure.

**GPT-4:** Elacestrant should be taken with food, as it increases the AUC by 22% and Cmax by 42% compared to fasted conditions, improving gastrointestinal tolerability. The FDA recommends this administration method to minimize gastrointestinal adverse reactions. Tmax details were not provided in the text.

Figure 5: An example (NDA217639) shows ChatGPT skips the essential details of key metrics after the length-constrained prompt. However, GPT-4 is able to retain the required details and generate a more concise summary than ChatGPT.

**Which model performs better in the study?**

The performance of GPT-4 in this study surpasses ChatGPT from multiple perspectives. First, in the human evaluation, GPT-4 received more votes from PSG assessors than ChatGPT. Figure 5 provides an example illustrating GPT-4's ability to generate summaries that capture all required details while maintaining conciseness, which ChatGPT fails. Second, GPT-4 demonstrates better adherence to the given length constraints than ChatGPT. All the summaries generated by GPT-4 meet the requirement in the length-constraint prompt, while only 89% of the summaries generated by ChatGPT follow the constraint. Furthermore, regarding factual consistency evaluation, 85% of the summaries generated by GPT-4 received a human majority vote indicating they are consistent with the reference summary in the drug label. The instances where GPT-4 is not consistent with reference summaries are mainly due to missing required PK values.

**Can GPT-4 be used in automatic evaluation?**

Since the food effect summarization would be used to assist PSG assessment, it is imperative for FDA professionals to evaluate the quality of the summary, though the evaluation by human is quite laborious and expensive. In this work, we find that GPT-4 evaluation provides a cheap and reasonable alternative to human evaluation. When using GPT-4 to rate the performance of different models against the reference summary, there are a number of issues. First, there exists an inconsistency problem whereby there is no guarantee that an LLM will produce the identical output for the same input every time. A good practice to fix the problem is to set temperature = 0; the temperature parameter of the model controls the randomness of the text generated, with 0 being deterministic. Second, we find a strong ordering effect with GPT-4 assigning higher scores to the model appearing earlier in the prompt. To mitigate such recency bias, we recommend reporting the mean score over both orders. Third, we observe from Figure 4 that GPT-4 assigns significantly higher scores to its own outputs compared to human ratings (64% vs. 43%), which represents the propensity of the model to favor its own suggestions. Future work should investigate the existence of potential biases in GPT-4 evaluation as well as possible mitigation strategies.

Despite the potential of GPT-4 for automating evaluation, it is not as flexible as humans, particularly in situations where specific definitions of summary quality are not provided. The performance of GPT-4 hence may vary due to the lack of clarity in evaluation standards. GPT-4 aligns with the human evaluation, concluding that it generates better summaries than ChatGPT. However, GPT-4 encounters difficulty distinguishing scenarios where both models produce equally good summaries. In cases where human assessors perceive the summaries as equally good, GPT-4 may favor a summary that includes relevant contextual information does not present in the other, such as "*explicitly linking it to a high-fat, high-calorie meal*" or "*mention the lack of clinically significant differences*". It also may penalize certain aspects based on subjective criteria, such as "*less concise*".

The factual consistency evaluation of GPT-4 shows a lower percentage than the human evaluation. There are two factors that could contribute to this discrepancy. First, possible bias exists in the human evaluation since the assessors serve as co-authors of the study despite being blinded to the source of a summary, which could have biased their assessments. Second, humans can adjust and reconcile minor numerical differences more flexibly than GPT-4. For example, while GPT-4 evaluates a summary that "*a high-fat meal affects its absorption, causing a 33% decrease in Cmax, a **10%** decrease in AUC*" which is not consistent with the reference summary stating "*a high-fat meal resulted in **9%** decrease in AUC and 33% decrease in Cmax compared to the fasted state*". Human assessors may consider the differences of 10% and 9% sufficiently close for factual consistency.

GPT-4 demonstrates an ability to understand the general idea presented in the summary. For example, it correctly identifies that "*the AUC and Cmax of asciminib significantly **decreased** when taken with food, with **greater reductions** observed for high-fat meals*" is consistent with "*The AUC and Cmax of asciminib decreased by 62% and 68%, respectively, with a high-fat meal (1000 calories, 50% fat) and by 30% and 35%, respectively, with a low-fat meal (400 calories, 25% fat).*" However, human assessors may have different standard for factual consistency, and expect the summary to include all the numerical values.

**Is automatic metrics effective at evaluation?**

Multiple studies have shown that automated metrics in NLP do not consistently align with human judgments, as they may not fully capture coherent sentence structure and semantics (Goyal et al., 2022; Zhang et al., 2023). Figure 6 shows the low correlation between the number of assessor votes and the ROUGE-L. Specifically, the Kendall Tau correlation coefficients for the "Most Preferred Summary" and "Least Preferred Summary" are 0.24 and -0.25, respectively. Moreover, the human evaluation results indicate that superior performance of GPT-4 is not fully captured by the ROUGE scores presented in Table 1, as it slightly lags behind ChatGPT in Turn 2 and 3. This finding reveals the limitations of relying solely on automatic metrics when applied to LLMs, an observation also consistent with previous research (Goyal et al., 2022).

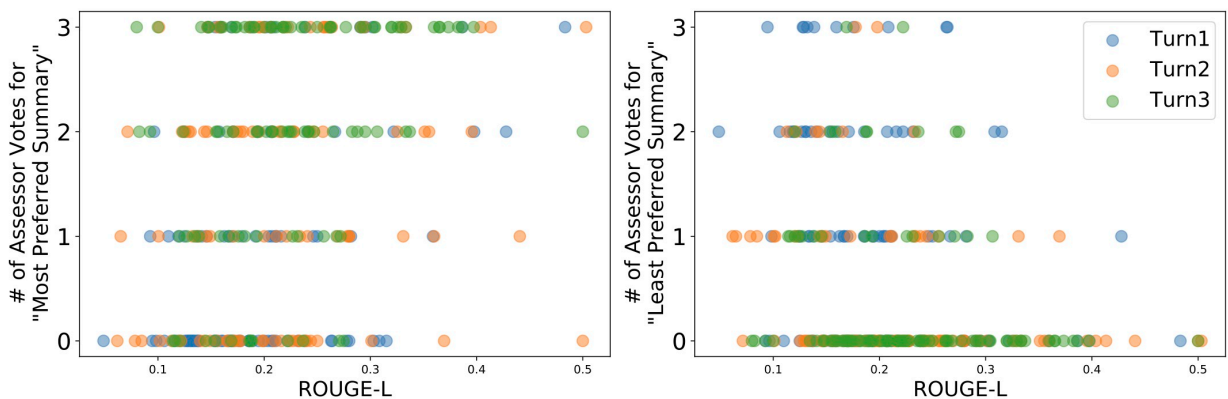

Figure 6: Correlation between the human evaluation score (the number of assessor votes) and ROUGE-L.

In conclusion, we propose a powerful strategy called iterative prompting, which enables human collaboration with the underlying LLM through multi-turn interaction. Extensive evaluations from both human raters and GPT-4 show that our approach has the capability to improve the quality of food effect summarization. These results suggest that GPT-4, as an AI assistant, may be able to aid in drafting food effect summaries that could be reviewed by human experts, thereby enhancing PSG assessment, and expediting the development of generic drug products. It is important to note that GPT-4 can be used to augment, rather than replace, human in the process of PSG assessment.

**Acknowledgments**: This work was partly supported by The United States Food and Drug Administration Contract #: 75F40119C10106

**Conflict of Interest:** The authors declare that they have no conflict of interest.

**Disclaimer:** The opinions expressed in this article are the author's own and do not reflect the view of the Food and Drug Administration, the Department of Health and Human Services, or the United States government.

**Author Contributions:** Conceptualization: Y.S. and H.L.; Methodology: Y.S. and H.L.; Formal analysis and investigation: Y.S. and H.L.; Writing – Original Draft, Y.S. and H.L.; Writing – Review & Editing, Y.S., M.H., T.V., F.A., L.Z., and H.L.; Funding Acquisition: H.L.; Resources: M.H., P.R., J.W., B.H., Y.Z. and L.Z.; Supervision: H.L.

# Appendix A. Summary of the Drug Products in the Study

| Clinical category | No. of Drug product | Drug substance | Dosage form |
|---|---|---|---|
| Immunosuppressant | 9 | Abrocitinib, belumosudil mesylate, baricitinib, upadacitinib, diroximel fumarate, monomethyl fumarate, dexamethasone, ozanimod HCl, ponesimod | Capsule and tablet |
| Antineoplastics | 30 | Infigratinib phosphate, pirtobrutinib, asciminib HCl, adagrasib, belzutifan, mobocertinib succinate, belzutifan, mobocertinib succinate, selinexor, alpelisib, apalutamide, sotorasib, tepotinib HCl, futibatinib, umbralisib tosylate, zanubrutinib, tivozanib HCl, avapritinib, enzalutamide pralsetinib, pemigatinib, selpercatinib, tucatinib, selumetinib sulfate, ceritinib, tazemetostat hydrobromide, erdafitinib, darolutamide, fedratinib HCl, capmatinib HCl, elacestrant DiHCl, azacitidine | Capsule and tablet |
| Anti-psychotics | 6 | Dexmethylphenidate HCl; serdexmethylphenidate chloride, lorazepam, amphetamine sulfate, calcium oxybate; magnesium oxybate; potassium oxybate; sodium oxybate, venlafaxine besylate, sertraline HCl | FDC capsule, capsule, and tablet |
| Antidiabetics | 3 | Finerenone, bexagliflozin, empagliflozin; linagliptin; metformin HCl | Tablet and ER tablet |
| Anti-inflammatory | 4 | Omaveloxolone, budesonide, celecoxib; tramadol HCl, celecoxib | FDC tablet, capsule, tablet, and oral solution |
| Anti-infection | 6 | Lefamulin acetate, amoxicillin; clarithromycin; vonoprazan fumarate, abacavir sulfate; dolutegravir sodium; lamivudine, Posaconazole, ibrexafungerp citrate, fexinidazole | Capsule, tablet, FDC capsule, FDC tablet, and for suspension |
| Anti-migraine | 4 | Rimegepant sulfate, ubrogepant, lasmiditan succinate, atogepant | Tablet and disintegrating tablet |
| Antihyperlipidemic | 4 | Atorvastatin calcium, bempedoic acid; ezetimibe, ezetimibe; rosuvastatin calcium, bempedoic acid | Suspension, FDC tablet, and tablet |
| Anticonvulsants | 4 | Topiramate, ganaxolone, cenobamate, fenfluramine HCl | Solution, suspension, and tablet |
| Androgenic | 5* | Tadalafil, finasteride; tadalafil, testosterone undecanoate, relugolix | Suspension, FDC capsule, and tablet |
| Contraception | 2 | Estradiol; norethindrone acetate; relugolix, drospirenone | FDC tablet and chewable tablet |
| Anticoagulant | 2 | Rivaroxaban, dabigatran etexilate mesylate | Suspension and pellets |
| Miscellaneous (e.g. relaxant, antianemia, osmotic laxative, and iron chelator and so on) | 21* | Ranolazine, baclofen, mavacamten, carbidopa; levodopa, mirabegron, levoketoconazole, sodium phenylbutyrate, sodium phenylbutyrate; taurursodiol, deferiprone, berotralstat HCl, lemborexant, lactitol, octreotide acetate, vibegron, solifenacin succinate, tramadol HCl, amlodipine benzoate, voxelotor, mitapivat sulfate | ER granules, granules, FDC tablet, ER suspension, capsule, pellets, suspension, tablets, DR capsule, solution, and FDC suspension |

*The same drug substance with different dosage forms
ER: extensive release, FDC: fix dosage combination, DR: delay release

**Appendix B. Prompt for GPT-4 Evaluation in Comparison between ChatGPT and GPT-4**

[System]

Please evaluate the performance of two automatically generated summaries of food effect study related text and determine which summary provides the most accurate and concise information, similar to the reference summary provided below.

Begin by stating either "A won", "B won", or "Tie" on a single line. Following that, please provide an impartial and comprehensive explanation of your evaluation, taking care to avoid any potential biases and ensuring that the order in which the summaries are presented does not influence your judgment.

[Reference Summary]

Coadministration of CIBINQO with a high-fat, high-calorie meal (total 916 calories, with approximate distribution of 55% fat, 29% carbohydrates, and 16% protein) had no clinically relevant effect on abrocitinib exposures (AUC and Cmax of abrocitinib increased by approximately 26% and 29%, respectively, and Tmax was prolonged by 2 hours).

[The Start of Summary A]

Abrocitinib, a JAK 1 inhibitor for treating atopic dermatitis, shows an increase in AUC by 26%, Cmax by 29%, and a 2-hour delay in Tmax when taken with a high-fat, high-calorie meal. However, these changes are not clinically significant. As a result, abrocitinib can be administered with or without food, as reflected in Phase 3 clinical trial designs.

[The End of Summary A]

[The Start of Summary B]

A food effect study on Abrocitinib found that taking the medication with a high-fat, high-calorie meal delayed Tmax by about 2 hours and increased the AUCinf and Cmax by 26% and 29%, respectively. Nonetheless, Abrocitinib can still be taken with or without food. In the pivotal bioequivalence study, the TBM formulation met bioequivalence criteria, showing similar median plasma Abrocitinib concentration-time profiles compared to the Phase 3 formulation under fasted conditions.

[The End of Summary B]

**Appendix C. Prompt for GPT-4 Evaluation in GPT-4 Consistency Evaluation**

[Reference Summary]

Coadministration of CIBINQO with a high-fat, high-calorie meal (total 916 calories, with approximate distribution of 55% fat, 29% carbohydrates, and 16% protein) had no clinically relevant effect on abrocitinib exposures (AUC and Cmax of abrocitinib increased by approximately 26% and 29%, respectively, and Tmax was prolonged by 2 hours).

[Model-generated Summary]

Abrocitinib, a JAK 1 inhibitor for treating atopic dermatitis, shows an increase in AUC by 26%, Cmax by 29%, and a 2-hour delay in Tmax when taken with a high-fat, high-calorie meal. However, these changes are not clinically significant. As a result, abrocitinib can be administered with or without food, as reflected in Phase 3 clinical trial designs.

[System]

To evaluate the model-generated summary of food effect study related text, please determine if it is factually consistent with the reference summary provided above.

Your evaluation should begin with a one-line answer of either "Yes" or "No". After that, please provide an unbiased and comprehensive explanation of your evaluation, taking care to avoid any potential biases.